\documentclass{article}

\usepackage{microtype}
\usepackage{graphicx}
\usepackage{subfigure}
\usepackage{booktabs} 

\usepackage{hyperref}
\usepackage{url}
\usepackage{booktabs}
\usepackage{amsfonts}
\usepackage{nicefrac}
\usepackage{microtype}
\usepackage{graphicx}
\usepackage{xcolor}
\usepackage{algpseudocode}
\usepackage{amsmath}
\usepackage{placeins}

\usepackage[accepted]{icml2025}

\usepackage{amsmath}
\usepackage{amssymb}
\usepackage{mathtools}
\usepackage{amsthm}

\usepackage[capitalize,noabbrev]{cleveref}

\theoremstyle{plain}

\theoremstyle{definition}

\theoremstyle{remark}

\usepackage[disable,textsize=tiny]{todonotes}

\icmltitlerunning{Spectrally Targeted Muon}

\begin{document}

\twocolumn[
\icmltitle{Spectrally Targeted Muon}



\icmlsetsymbol{equal}{*}

\begin{icmlauthorlist}
\icmlauthor{Vishrut Goyal}{stanfordcs}
\icmlauthor{Rohan Ramkumar}{stanfordee}

\end{icmlauthorlist}

\icmlaffiliation{stanfordcs}{Department of Computer Science, Stanford University}
\icmlaffiliation{stanfordee}{Department of Electrical Engineering, Stanford University}

\icmlcorrespondingauthor{Vishrut Goyal}{vishrutg@stanford.edu}
\icmlcorrespondingauthor{Rohan Ramkumar}{rohansr@stanford.edu}

\icmlkeywords{Muon, orthogonalized updates, Newton-Schulz, effective rank}

\vskip 0.3in
]



\printAffiliationsAndNotice{} 
\begin{abstract}
    The Muon optimizer orthogonalizes each update matrix, setting all of its singular values to one, and has proven highly effective for training large language models. It remains unclear, however, whether this success comes from amplifying small singular directions that gradient descent neglects or from suppressing large, degenerate directions that disrupt training. We introduce Spectrally Targeted Muon, which orthogonalizes only the singular values above or below a threshold $\tau$, so that varying $\tau$ interpolates between normalized SGD and Muon. It isolates the relevant singular subspaces with projections computed by Newton-Schulz iteration on a shifted Gram matrix, so no SVD is needed. We evaluate these variants on the CIFAR-10 and NanoGPT speedruns, tracking the effective rank of gradient, update, and weight matrices and a new metric, the alignment of updates with the tangent space of the weight matrix's isospectral manifold. We find three things. First, the small singular values of the momentum are not noise. Orthogonalizing everything except the few largest singular values of each matrix nearly matches Muon while touching only a small fraction of the momentum, whereas orthogonalizing only the top falls well short even though it holds almost all of it. On language models every momentum singular value is far below one, so targeted orthogonalization can only amplify, and Muon wins by making directions that are too small to train on at their raw scale trainable. Second, shrinking the largest singular values is what keeps the parameter spectrum flat; this is cheap, and it is not what drives the loss. Third, AdamW differs from Muon mainly in how slowly it builds structure, which explains its slower start and why warmup helps AdamW but only hurts Muon.

\end{abstract}
\section{Introduction}
The Muon optimizer for deep learning works by orthogonalizing the update matrix \citep{jordan2024muon}. Unlike many other proposed optimizers, this has been highly successful empirically for training large language models, especially when coupled with weight decay \citep{liu2025muonscalablellmtraining}. This is because the matrix can be orthogonalized relatively efficiently, using methods such as Newton-Schulz or Polar Express \citep{amsel2026polarexpressoptimalmatrix}.

The orthogonalization step can be thought of as applying the matrix sign function to the gradient matrix, which normalizes all the singular values to one, prompting the question of whether Muon's success comes from amplifying small but important directions with small singular values ignored by standard gradient descent, or from normalizing large, degenerate directions with large singular values that misdirect the training.

In this paper, we will explore a method to target this orthogonalization to singular values either above or below a certain threshold $\tau$, and we will also experiment with different values for this threshold, thus allowing us to interpolate between SGD and Muon with a spectrally targeted version. Specifically, we will apply these Muon variants to the two speedrun tasks originally used to show Muon's performance, the CIFAR-10 dataset with a convolutional network, and the NanoGPT large language model, just as in \citet{jordan2024muon}. In addition to standard performance metrics, we will examine other metrics more pertinent to the linear-algebraic structure of our weight and update matrices. This allows us to explore what characteristics of matrix optimizers lead to richer learning.
\section{Related Work}
Recent Muon variants like NorMuon and Aurora mitigate neural death in Muon via sophisticated row-wise (neuron) normalization \citep{li2025normuonmakingmuonefficient, dewulf2026aurora}. In contrast, our approach normalizes singular values to explicitly target the weights' underlying linear structure.

While state-of-the-art orthogonalization uses the adaptive Polar Express algorithm \citep{amsel2026polarexpressoptimalmatrix}, we instead build on Newton-Schulz iterations using both the speedrun polynomial which yields singular values distributed as $\mathrm{Uniform}(0.5, 1.5)$ and the convergent polynomial which iterates to the exact matrix sign function \citep{jordan2024muon}.

Other works also attempt to reshape the singular value distribution to something in between the raw momentum distribution and the flat distribution from Muon. While MuCon and spectral gradient clipping bound singular values \citep{yi2026muconclippedmuonupdates,yukhimchuk2026gradientclippingvectornorms} at some threshold, our method discontinuously forces them to one based on the $\tau$ threshold. This isolates whether Muon's success stems from amplifying minor components or suppressing dominant ones by applying full Muon in one subspace and raw SGD in another. Furthermore, our approach relies entirely on efficient matrix multiplications, bypassing costly SVD computations. Several works also reshape Muon's update spectrum with a continuous transform instead \citep{schattenp2026,muonp2026,dynmuon2026,pion_highpass2026,pion_preserving2026}. In particular, \citet{freon2026} smoothly interpolates between Muon and SGD by varying decay terms. That work also shows that randomly set singular values can perform almost comparably to Muon. This motivates our question of which shifts in the spectrum relative to Muon have actual impact.

\section{Methods}
The fundamental algorithm underlying Muon's orthogonalization step is the Newton-Schulz algorithm for calculating the matrix sign function. For an $n\times n$ diagonal matrix $\Sigma,$ with diagonal elements $\sigma_1,\dots,\sigma_n,$ the matrix sign function $\mathrm{sign}$ applied to $\Sigma$ is given by the $n\times n$ diagonal matrix with entries $\mathrm{sign}(\sigma_1),\dots,\mathrm{sign}(\sigma_n),$ which is orthogonal.

Given an arbitrary matrix $M$ with (reduced/thin) singular value decomposition $U\Sigma V^\top,$ we define \[\mathrm{sign}(M)=\mathrm{sign}(U\Sigma V^
\top)=U\mathrm{sign}(\Sigma)V^\top=UV^\top,\] as $\Sigma$ consists of positive singular values on the diagonal and thus $\mathrm{sign}(\Sigma)=I.$
\subsection{Classical Newton-Schulz}
The standard method computes this via polynomial iteration. Odd polynomials like $p(x) = 3x - \frac{16}{5}x^3 + \frac{6}{5}x^5$ asymptotically converge to $\mathrm{sign}(x)$ on the interval $\left[-1, 1\right]$. 

Extended to matrices, the polynomial acts like
\[ p(M) = 3M - \frac{16}{5}(MM^\top)M + \frac{6}{5}(MM^\top)^2M \]
Iterating $p(M)$ converges to $\mathrm{sign}(M)$. To guarantee convergence, $M$ must first be normalized by its Frobenius norm $\sqrt{\mathrm{tr}(M^\top M)}$, which ensures all singular values are bounded above by 1. 

For performance, $p(M)$ is usually iterated only a few times. As a result, alternative polynomials are often chosen to optimize short-term convergence rather than perfect asymptotic convergence, and as Frobenius scaling shrinks the singular values of poorly conditioned matrices close to 0, their early iterative behavior is heavily dominated by the polynomial's linear term. 

\subsection{Spectrally Targeted Newton-Schulz}

To selectively orthogonalize singular values separated by a threshold $\tau>0$, we isolate the relevant subspaces using projection matrices. Consider the shifted symmetric matrix $A=M^\top M-\tau^2 I$. The eigenvalues of $A$ are the squared singular values of $M$ shifted down by $\tau^2$; thus, its negative and positive eigenvalues neatly partition the singular values of $M$ below and above $\tau$.

By computing $S=\mathrm{sign}(A)$, we can define the projection matrices for the bottom and top singular subspaces as $P^\mathrm{bot}=\frac{1}{2}(I-S)$ and $P^{\mathrm{top}}=\frac{1}{2}(I+S)$. Finally, we selectively apply Newton-Schulz orthogonalization to the desired subspace by calculating either $\mathrm{sign}(M)P^{\mathrm{top}}+MP^\mathrm{bot}$ (targeting top) or $MP^{\mathrm{top}} + \mathrm{sign}(M)P^{\mathrm{bot}}$ (targeting bottom), as summarized in Algorithm \ref{alg:example}. Since this algorithm requires orthogonalizing $M_t^\top M_t-\tau^2 I$, which can be far more poorly conditioned than $M$, for NanoGPT we use the convergent polynomial for both Newton-Schulz calls, iterated 20 times in fp32, so that the sign is resolved even near $\tau$. In the training runs the lifted singular values land within $0.1\%$ of one, and the number lifted matches the number on the targeted side of $\tau$ to within one per matrix. Since $\tau$ is absolute, a matrix with no singular value above $\tau$ just receives its raw momentum.

\begin{algorithm}[tb]
   \caption{Spectrally Targeted Muon}
   \label{alg:example}
\begin{algorithmic}[1]
\Require Learning rate $\eta$, momentum $\mu$, threshold $\tau$
\State Initialize $B_0 \gets 0$
\For{$t = 1, \dots$}
    \State Compute gradient $G_t \gets \nabla_\theta \mathcal{L}_t(\theta_{t-1})$
    \State $B_t \gets \mu B_{t-1} + G_t$
    \State $M_t \gets G_t + \mu B_t$ \Comment{Nesterov momentum}
    \State $S_t\gets \text{NewtonSchulz}(M_t^\top M_t-\tau^2 I)$
    \State $P^{\mathrm{top}}_t\gets\frac{1}{2}(I+S_t)$, $P^{\mathrm{bot}}_t\gets\frac{1}{2}(I-S_t)$
    \If{Targeting Top}
    \State $O_t \gets \text{NewtonSchulz}(M_t)P^{\mathrm{top}}_t + M_t P^{\mathrm{bot}}_t$
    \ElsIf{Targeting Bottom}
    \State $O_t \gets  M_t P^{\mathrm{top}}_t + \text{NewtonSchulz}(M_t)P^{\mathrm{bot}}_t $
    \EndIf
    \State Update parameters $\theta_t \gets \theta_{t-1} - \eta O_t$
\EndFor
\State \textbf{return} $\theta_t$
\end{algorithmic}
\end{algorithm}
\subsection{Evaluation}\label{tangent}
We used two special metrics to study the linear-algebraic structure of the learning, namely effective rank and projection onto the tangent space of the isospectral manifold.

\subsubsection*{Effective Rank} The effective rank is defined as the exponent of the spectral entropy. That is, if you normalize the singular values $\{s_1,...,s_n\}$ by $\sum_{i=1}^n s_i$ to get probabilities $\{p_1,...,p_n\}$, the effective rank is $\exp(-\sum _{i=1}^np_i\log p_i)$. Similar to perplexity, this number can be interpreted as the effective number of equally likely directions a random vector can be mapped to \citep{7098875}. 
\subsubsection*{Isospectral Projection}
Muon orthogonalizes the update matrix, but how this affects the spectrum of the
parameter matrix is unclear. Random initialization yields approximately
dynamically isometric weight matrices, whose singular values are
well-conditioned \citep{pennington2017resurrectingsigmoiddeeplearning}, so
training must move the parameters away from this state, and an optimizer's
influence on the spectrum determines how much it does so. Consider a weight matrix $W$, and let $\mathcal M$ be its isospectral manifold, consisting of matrices with the same singular values as $W.$ Given an update $Z$, we measure $\rho(Z;W),$ which is the fraction of $Z$ that lies in $T_W\mathcal M,$ the tangent space of $\mathcal M$ at $W$ (see \hyperref[appendix]{Appendix} for a derivation or $\rho$).
 A value of
$\rho$ near one means the update mostly rotates the singular bases of $W$ while
preserving its spectrum, and a smaller value means more of the update changes the
singular values. In practice $\rho$ is very close to one, so we report the
log-barrier $-\log(1-\rho)$ to make small differences across training steps and
optimizers visible.

\section{Results and Discussion}

\subsection{CIFAR-10 Training}

We train a six-layer convolutional neural network on the CIFAR-10 dataset, building off of Keller Jordan's original speedrun using Muon \citep{jordan202494}. We chose this approach because the speedrun ensures exhaustively tuned learning rates and architectures, making it less likely that our results are artifacts of hyperparameters. This dataset includes 50,000 training images divided into 10 classes, along with 10,000 images for validation \citep{krizhevsky2009learning}. 
\begin{figure}
\centering
\includegraphics[width=\columnwidth]{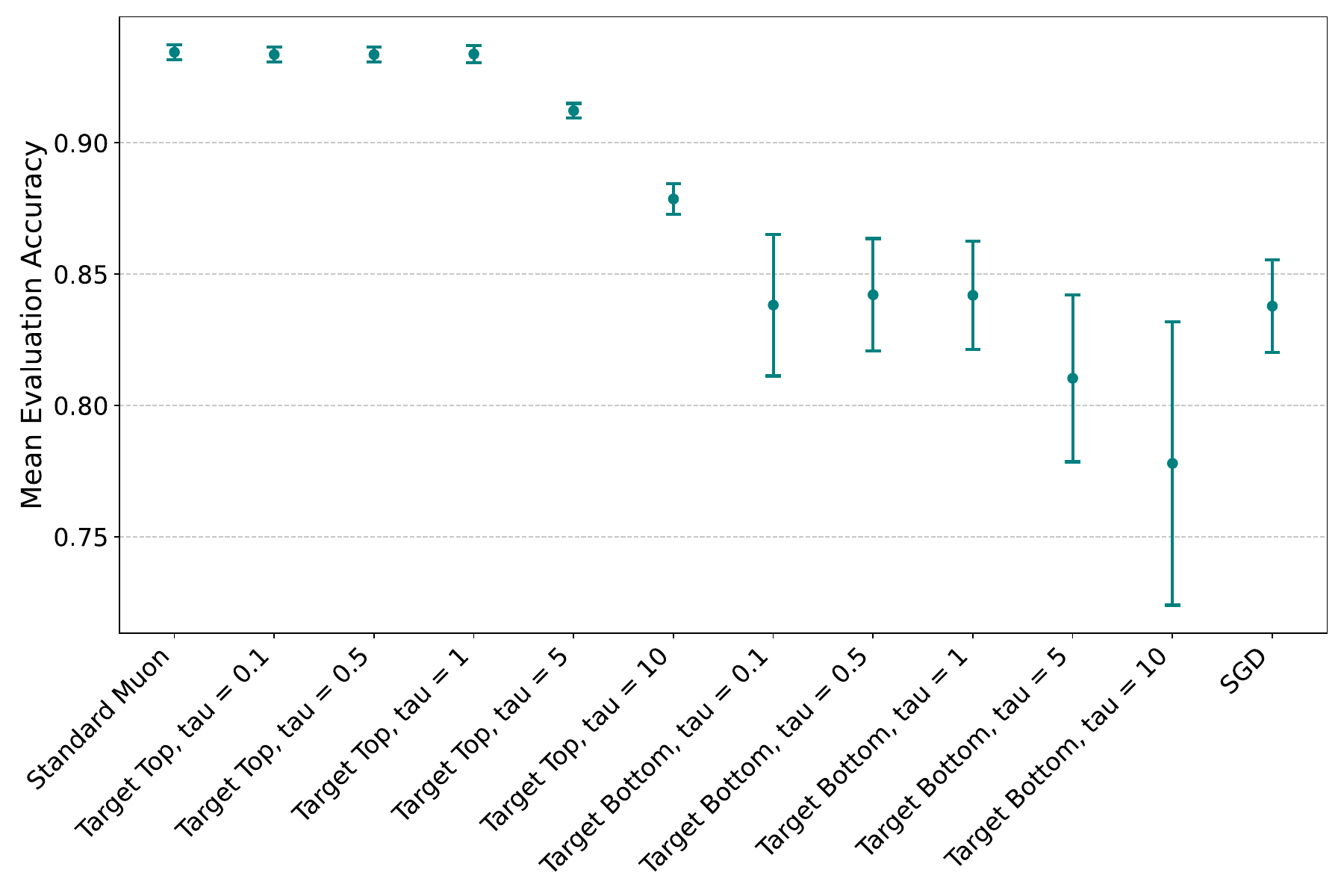}
\caption{Mean evaluation accuracy across 20 trials for each variant, with error bars representing $\pm$2 standard deviations away from the mean.}
\end{figure}
Validation accuracy was used as the primary metric for model performance. Learning rates were held constant across all optimizers, including SGD, and equal Nesterov momentum was applied. Instead of random flips for augmentation, all data is flipped every epoch, which leads to faster training. Our standard Muon run was therefore nearly identical to the speedrun implementation, aside from a batch size of 500 rather than 2000 due to GPU limitations. Therefore, we achieved an accuracy of 93.6\% as opposed to the 94\% in Jordan's paper. This aligns with existing work arguing that Muon takes particular advantage of large batch sizes in much larger models \citep{ai2025practicalefficiencymuonpretraining}. 

We also measured the effect of switching to the convergent Newton-Schulz polynomial, and increasing the number of iterations used from 3 to 7. For pure Muon, this left mean accuracy over 20 runs unchanged at 93.5\%. However, for top targeted Muon with $\tau=1$, it brought mean accuracy from 91.2\% to 93.4\%. This is because our method orthogonalizes something near $M^\top M$, which has a condition number squared compared to $M$, requiring more accuracy near 0. Thus, this setup was used for all CIFAR experiments. 

\begin{figure*}
\centering
\begin{tabular}{cc}
    \includegraphics[width=0.3\textwidth]{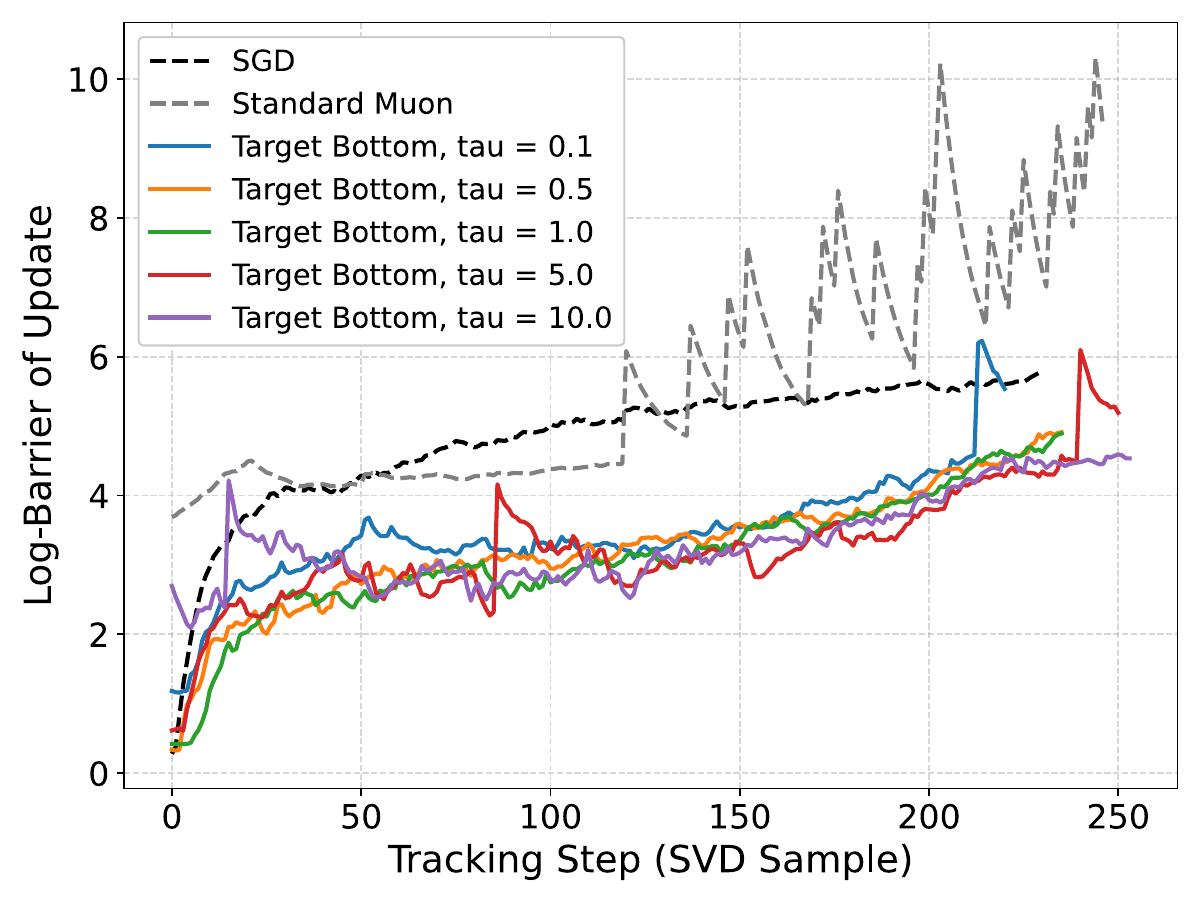}&
    \includegraphics[width=0.3\textwidth]{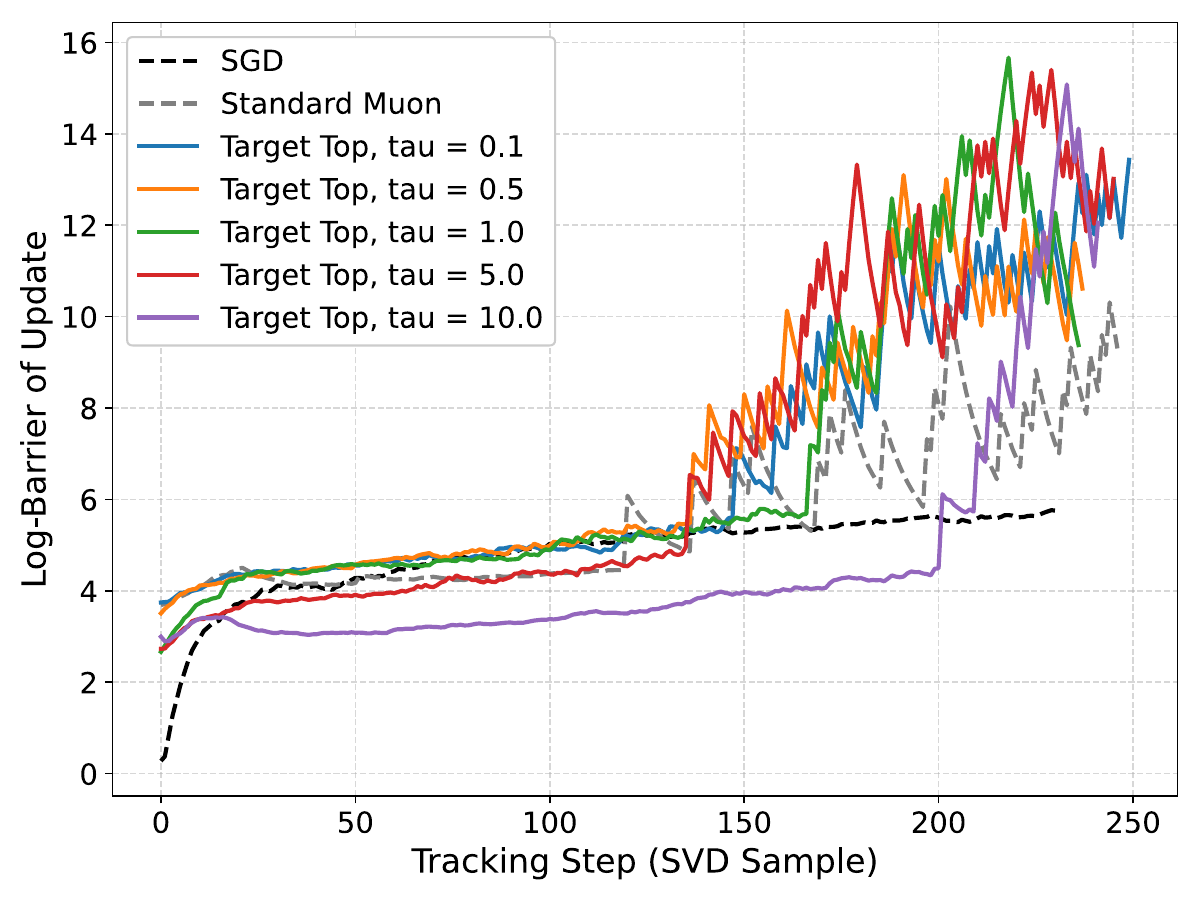}\\
    \begin{tabular}{@{}c@{}} (a) Log-barrier of tangent alignment \\for Muon targeted on bottom \end{tabular}&\begin{tabular}{@{}c@{}} (b) Log-barrier of tangent alignment \\for Muon targeted on top \end{tabular}\\
    \includegraphics[width=0.3\textwidth]{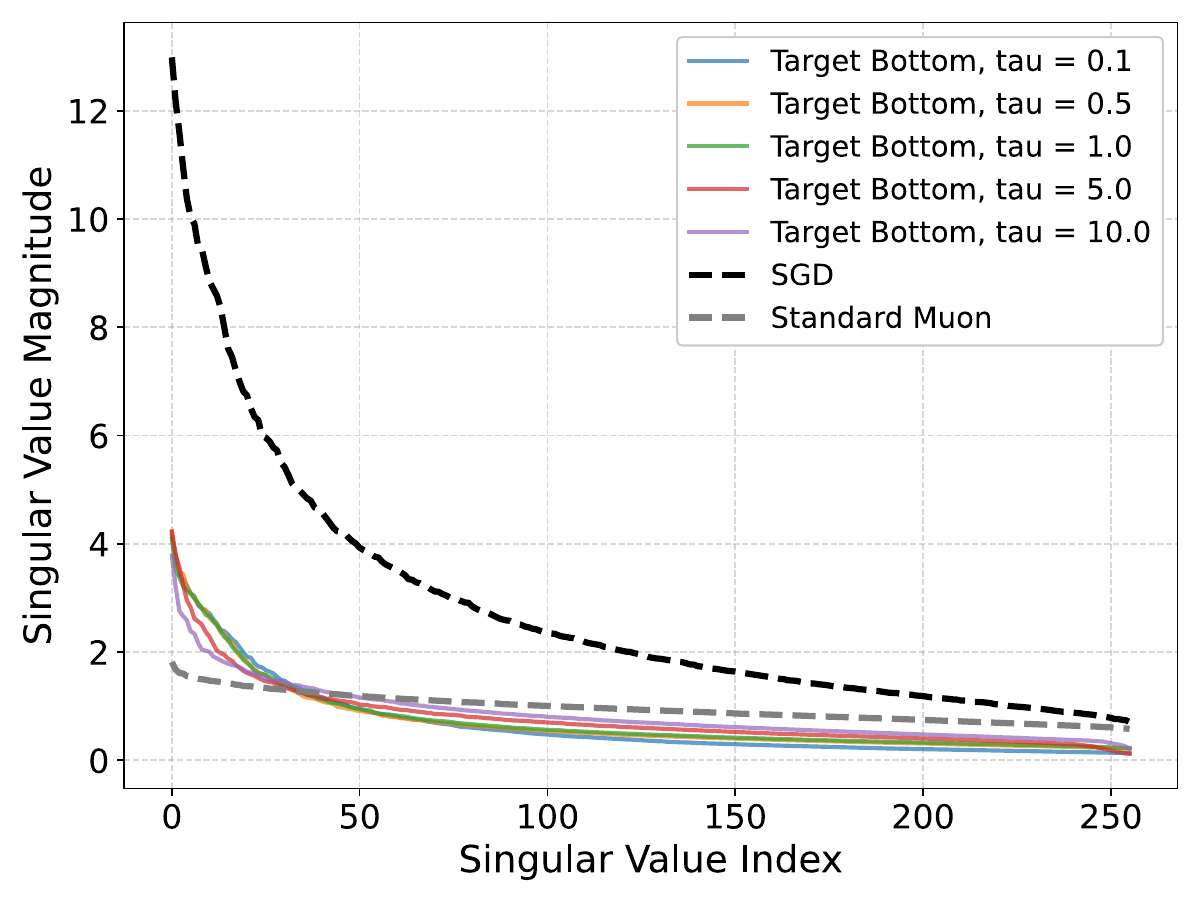}&
    \includegraphics[width=0.3\textwidth]{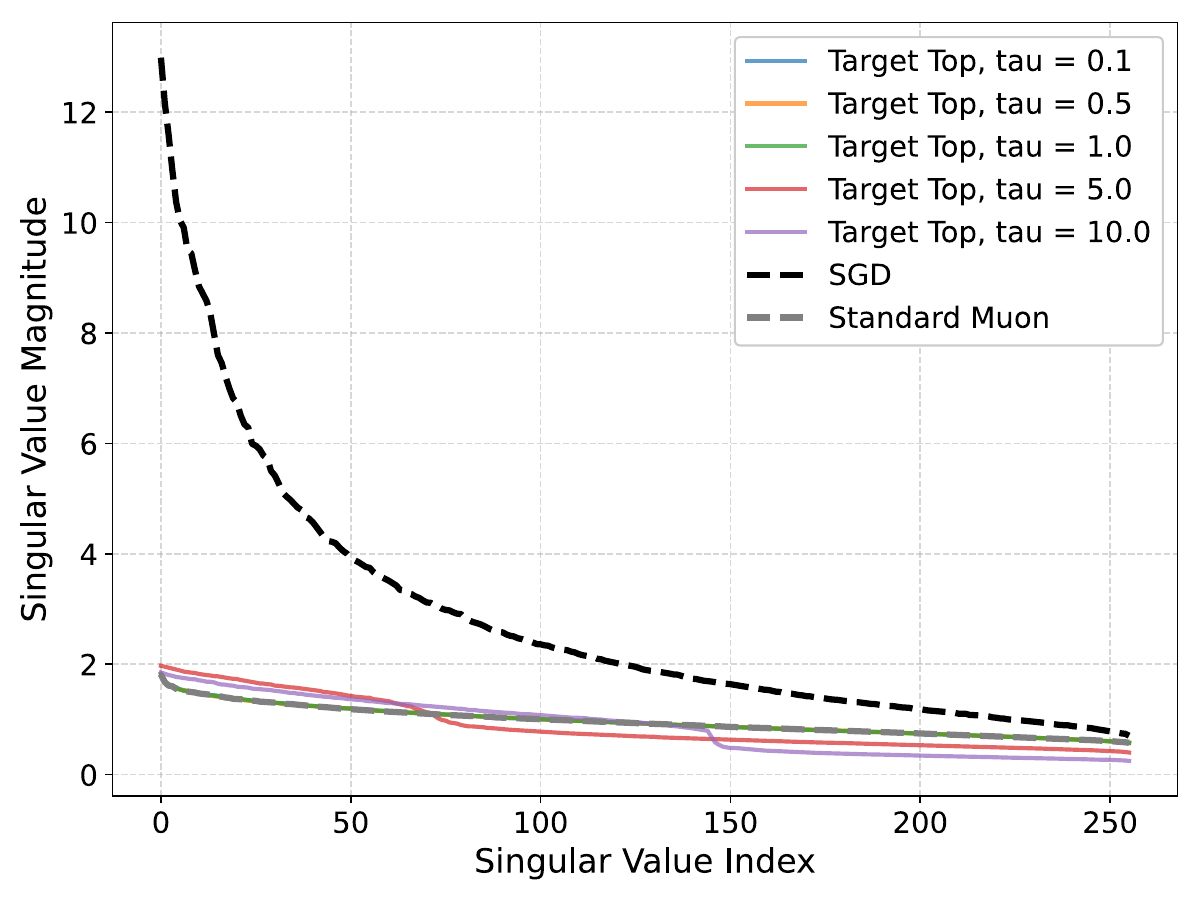}\\
    \begin{tabular}{@{}c@{}} (c) Final weight spectrum \\for Muon targeted on bottom \end{tabular}&\begin{tabular}{@{}c@{}} (d) Final weight spectrum \\for Muon targeted on top \end{tabular}\\
\end{tabular}
\caption{\label{tangentspectra} Tangent alignment and weight spectrum plots for CNN trained with bottom and top targeted Muon over 8 training epochs. }
\end{figure*}

Comparing different levels of top singular value thresholds, we see a smooth decline from the performance of Muon towards the performance of SGD. This includes a statistically significant drop from $\tau = 1$ to $\tau=5$, another significant drop from 5 to 10, and finally a drop from 10 to pure SGD with momentum. On the other side, $\tau \leq 1$ all had performance which was statistically indistinguishable from Muon. This aligns with our intuition that clipping at a value above the maximum singular value behaves like no orthogonalization at all, and clipping at a value below the majority of singular values will behave like full orthogonalization. Further support for this is provided by the spectrum of the final trained weight, which aligns nearly completely with the weight trained using Muon. Also, note that even clipping at high levels still produces very well-conditioned final weight matrices compared to SGD. Examining the tangent alignment, an interesting behavior is that alignment strongly increases after 150 to 200 steps, for all top clipped versions. This suggests that Muon's positive impact on condition comes largely from suppressing the largest gradient singular values.  

Looking at the tangent alignment in subplots (a) and (b) of Figure \ref{tangentspectra}, we see that targeting Muon on the bottom singular values consistently yields worse alignment than even standard SGD, while targeting the top clears Muon, especially toward the end of training. This fits with the observed performance, which is similar to raw SGD, and becomes worse as we clip at higher and higher levels. This is because our first experiment shows that the significant singular values are largely greater than 1. This means that clipping lower singular values to 1 makes the top singular value subspace make up a more disproportionate part of the update mass. This supports our hypothesis that the top singular value subspace is the least aligned with the isospectral tangent. Intuitively, we can think of this as reinforcing information already contained within the parameter, rather than learning new facts.

\subsection{NanoGPT Training}
We train a 12-layer transformer with 768-dimensional embeddings on 0.80B tokens of FineWeb, validating on 10.5M held-out tokens \citep{penedo2024finewebdatasetsdecantingweb}. Similar to CIFAR-10, we build on Jordan's modded-nanogpt speedrun \citep{jordan2024moddednanogpt}, which trains to a 3.28 validation loss in 1530 steps with Muon. Each transformer block consists of $W_Q$, $W_K$, and $W_V$ matrices which generate the query, key and value for multi-head self-attention, before passing through output matrix $W_O$. The MLP block contains two projection matrices in and out of the hidden space. In addition, there are skip connections connecting layers from the first and second half, similar to U-Net \citep{DBLP:journals/corr/RonnebergerFB15}. All output matrices were initialized to 0, making the model initialize to the identity function and allowing rapid gradient flow to layers during initialization. All runs use a linear learning rate cooldown over the last 600 steps. Muon and its variants use a learning rate of 0.05 with no warmup, except for one warmup control, and no rescaling between variants, and AdamW uses the speedrun's tuned learning rate of 0.0018 \citep{karpathy2022nanogpt}, with and without the canonical 250-step warmup. As the unorthogonalized endpoint we use NSGD, the Nesterov momentum divided by its spectral norm, at the same learning rate.\footnote{Our power-iteration estimate of the spectral norm fails on matrices with very small momentum (about 5--10\% of matrix-steps) and produces oversized updates there, so NSGD is an imperfect endpoint; a correct implementation would likely do at least as well.} We run both top and bottom targeting at $\tau\in\{10^{-4},2\times10^{-4},4\times10^{-4},10^{-3},3\times10^{-3}\}$ with three seeds each, three for Muon, two for the other optimizers, and one for the learning rate controls.

\begin{figure*}[t]
\centering
\begin{tabular}{cc}
    \includegraphics[width=0.4\textwidth]{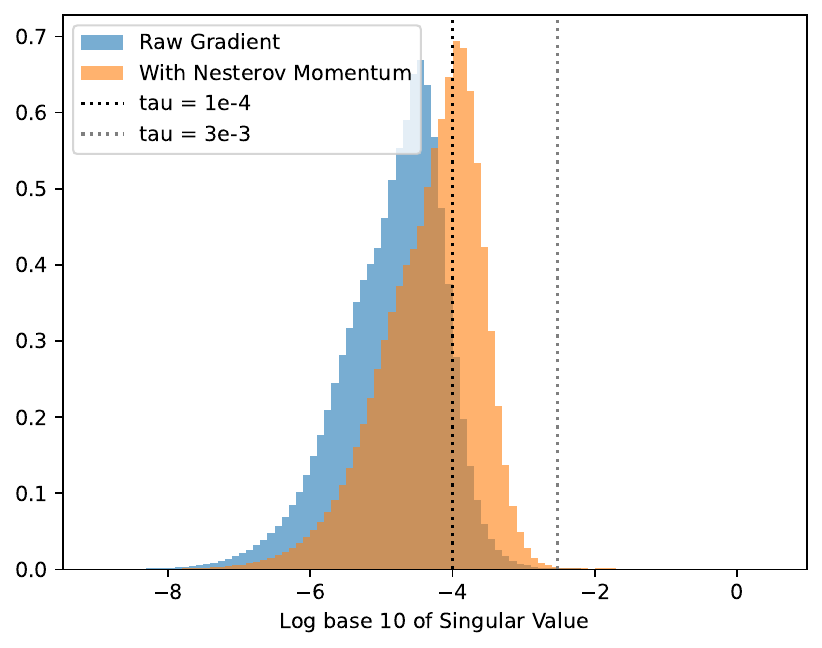} & \includegraphics[width=0.4\textwidth]{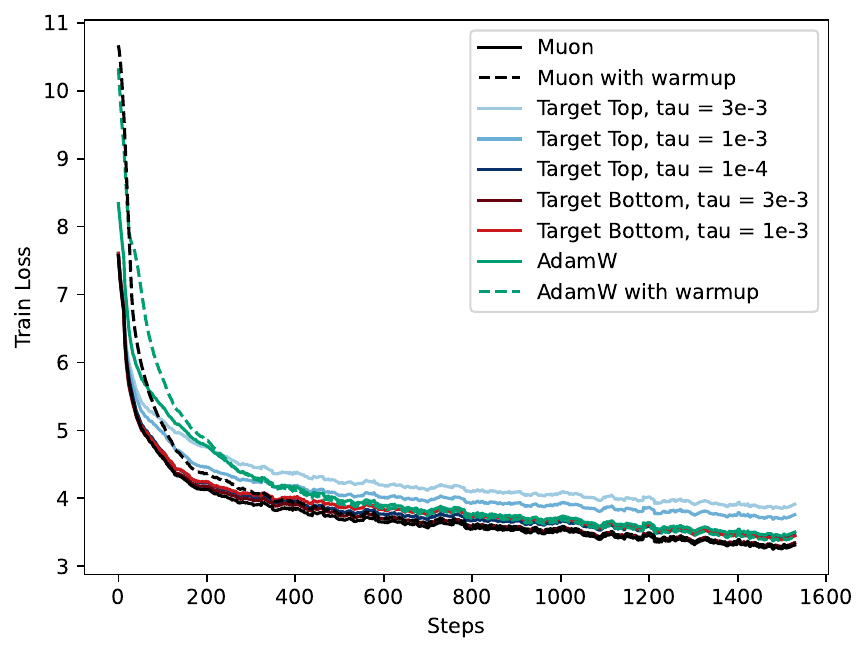} \\
    \parbox[t]{0.4\textwidth}{\centering (a) Gradient and momentum singular values} & \parbox[t]{0.4\textwidth}{\centering (b) Train loss comparison} \\
\end{tabular}
\begin{tabular}{cc}
    \includegraphics[width=0.4\textwidth]{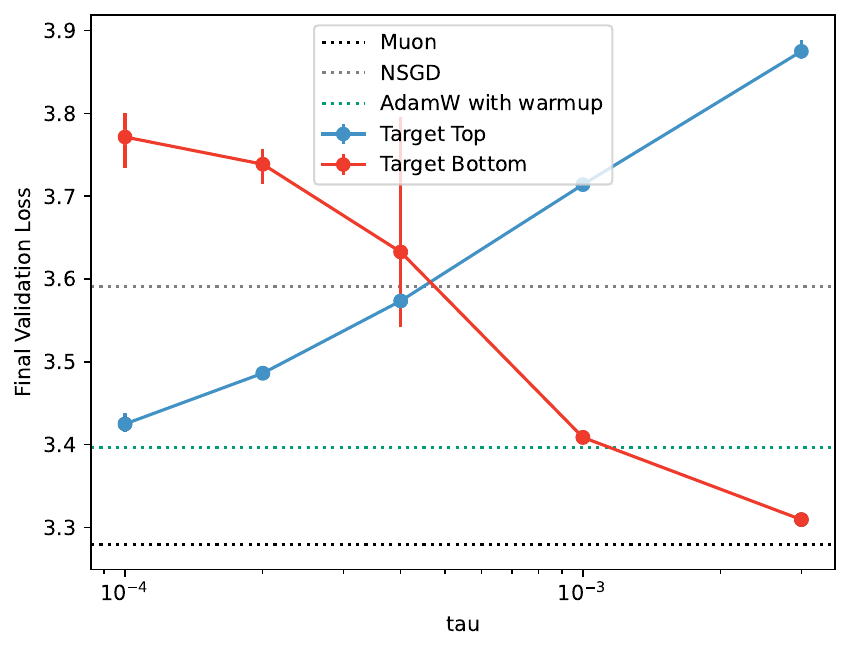} & \includegraphics[width=0.4\textwidth]{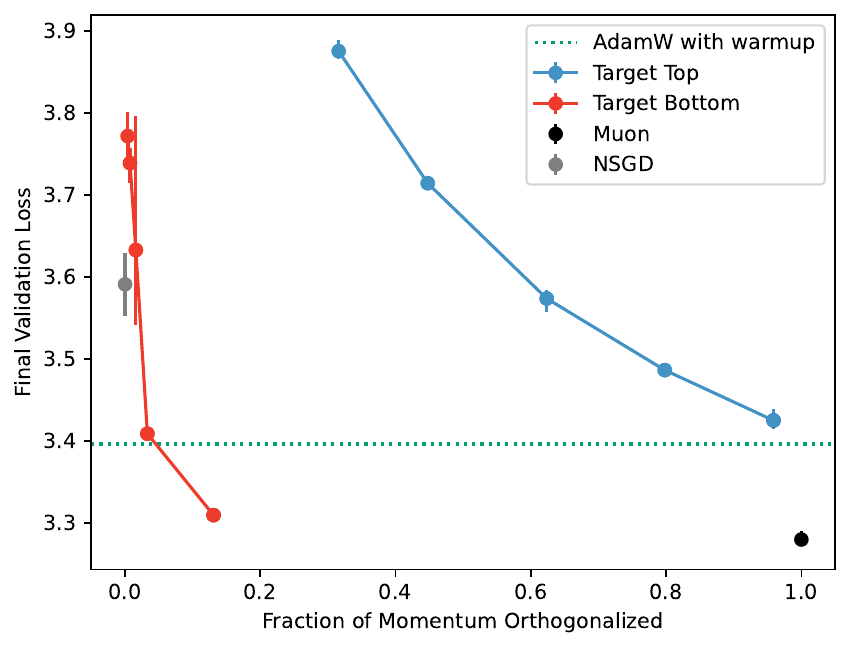} \\
    \parbox[t]{0.4\textwidth}{\centering (c) Final loss vs.\ threshold} & \parbox[t]{0.4\textwidth}{\centering (d) Final loss vs.\ fraction of momentum orthogonalized} \\
\end{tabular}
\vspace{-4mm}\caption{\label{distloss}Singular values of the raw gradient and of the momentum that is orthogonalized (a; dotted lines mark the smallest and largest $\tau$), training loss (b), and final validation loss against the threshold (c) and against the fraction of the momentum that is orthogonalized (d; by squared Frobenius norm, matrix average). Every momentum singular value lies far below one, so the targeted band is always amplified to one; Muon and NSGD are the endpoints of the family.}
\end{figure*}

\begin{figure*}[t]
\centering
\begin{tabular}{cc}
    \includegraphics[width=0.3\textwidth]{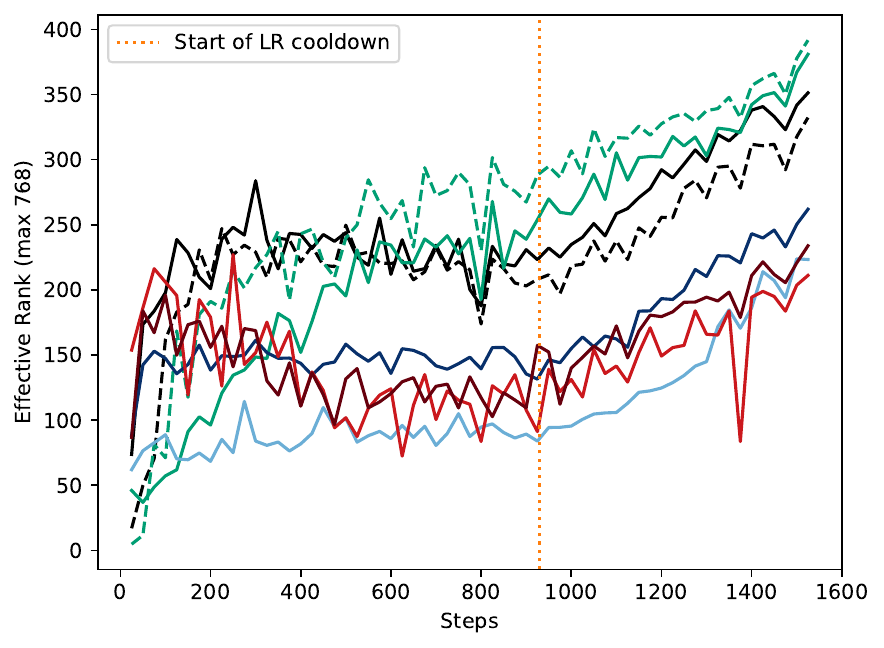} & \includegraphics[width=0.3\textwidth]{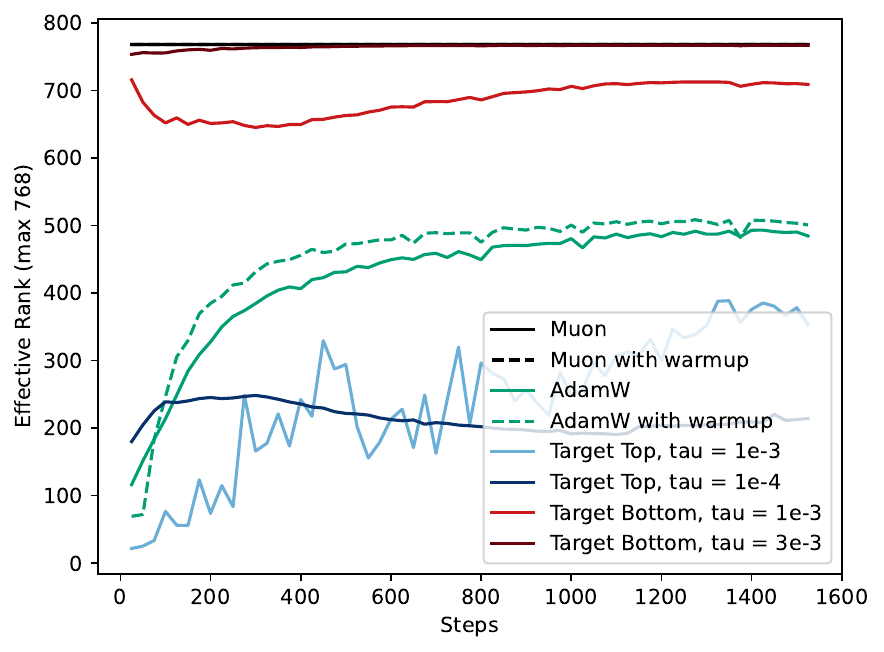} \\
    \parbox[t]{0.3\textwidth}{\centering (a) Gradient effective rank} & \parbox[t]{0.3\textwidth}{\centering (b) Update effective rank} \\
\end{tabular}
\begin{tabular}{ccc}
    \includegraphics[width=0.3\textwidth]{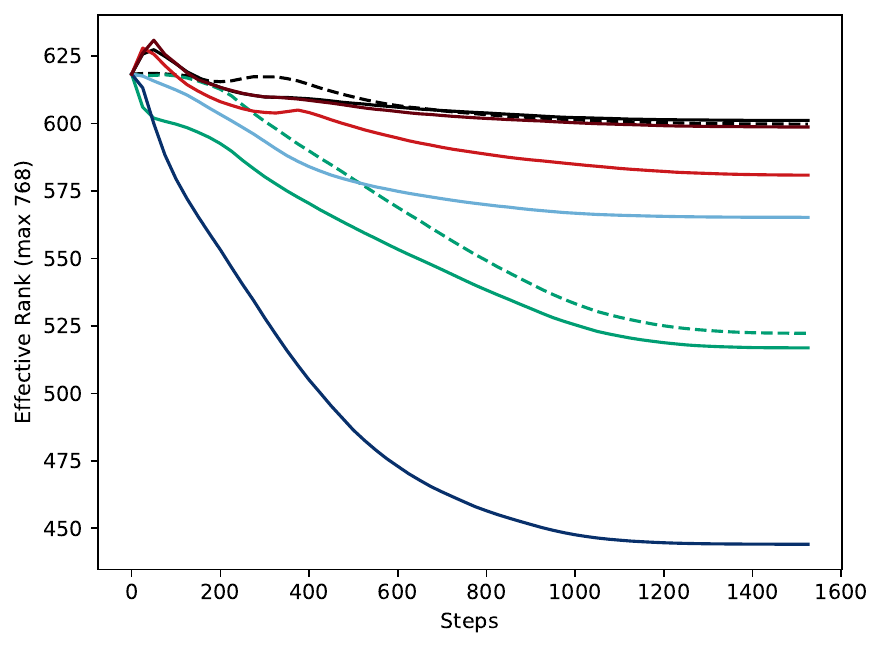} & \includegraphics[width=0.3\textwidth]{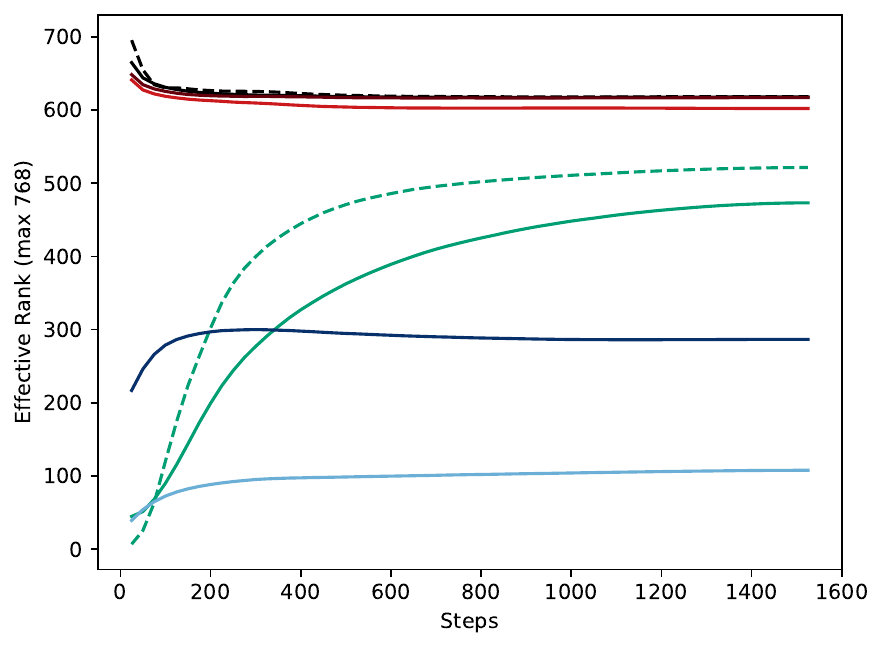} & \includegraphics[width=0.3\textwidth]{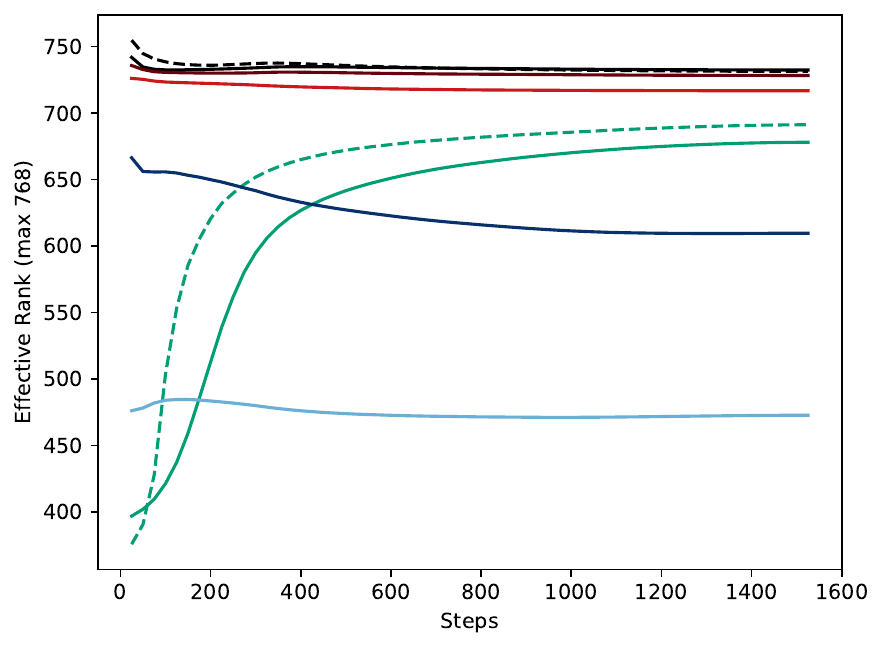} \\
    \parbox[t]{0.3\textwidth}{\centering (c) Value matrix effective rank} & \parbox[t]{0.3\textwidth}{\centering (d) Output matrix effective rank} & \parbox[t]{0.3\textwidth}{\centering (e) MLP effective rank} \\
\end{tabular}
\vspace{-4mm}\caption{\label{eranks}Effective rank of the gradient (a) and update (b), averaged over all matrices, and of the value (c), attention-output (d) and MLP (e) parameters, averaged over layers (seed 0). Dashed lines are the warmup variants, the dotted line in (a) marks the start of the learning rate cooldown, and the legend in (b) applies to all panels.}
\end{figure*}

\begin{table*}[ht]
\centering
\caption{Final FineWeb validation loss at step 1530, mean $\pm$ half-range over seeds (three seeds for the targeted family and Muon, two for the other optimizers, one for the learning rate controls). $\dagger$: at least one seed shows a loss spike; for the bottom arm these are carried by the MLP output matrices, which these thresholds never lift, when a hard batch near step 1020 passes through them un-normalized.}
\label{final_val_loss}
\resizebox{\textwidth}{!}{%
\begin{tabular}{lccccc}
\toprule
 & $\tau=3\times10^{-3}$ & $\tau=10^{-3}$ & $\tau=4\times10^{-4}$ & $\tau=2\times10^{-4}$ & $\tau=10^{-4}$ \\
\midrule
Top-targeted    & 3.875 $\pm$ .011 & 3.714 $\pm$ .002 & 3.573 $\pm$ .013 & 3.486 $\pm$ .004 & 3.425 $\pm$ .012 \\
Bottom-targeted & \textbf{3.309} $\pm$ .002 & 3.409 $\pm$ .003 & 3.633 $\pm$ .127$^\dagger$ & 3.739 $\pm$ .022$^\dagger$ & 3.772 $\pm$ .033$^\dagger$ \\
\midrule
\multicolumn{2}{l}{Muon} & \textbf{3.280} $\pm$ .008 & \multicolumn{2}{l}{AdamW w/ warmup} & 3.396 $\pm$ .004 \\
\multicolumn{2}{l}{Muon w/ warmup} & 3.309 $\pm$ .001 & \multicolumn{2}{l}{AdamW w/ warmup, wd 0} & 3.397 $\pm$ .004 \\
\multicolumn{2}{l}{Muon, lr 0.005} & 3.332 & \multicolumn{2}{l}{AdamW} & 3.468 $\pm$ .008 \\
\multicolumn{2}{l}{NSGD} & 3.591 $\pm$ .038$^\dagger$ & \multicolumn{2}{l}{Top $\tau=3\times10^{-3}$, lr 0.49} & 4.800 \\
\multicolumn{2}{l}{Top $\tau=10^{-3}$, lr 0.29} & 4.392 & \multicolumn{2}{l}{Top $\tau=10^{-4}$, lr 0.095} & 3.521 \\
\bottomrule
\end{tabular}%
}
\end{table*}
Figure~\ref{distloss}(a) shows that the singular values are spread across many orders of magnitude, indicating that gradient matrices in the language model setting are very poorly conditioned. This variation does not stem from variation between different matrices: as we can see in Figure~\ref{eranks}(a), the effective rank of the gradient stays below about 280 for most of training. This implies a low-entropy and highly nonuniform distribution within each matrix. Figure~\ref{distloss}(a) also displays the distribution of the gradients after Nesterov momentum is applied. Momentum thins out the low end of the spectrum, suggesting that it suppresses some noisy directions, but otherwise leaves its shape alone and simply shifts scale.

This distribution was used to select the absolute thresholds $\tau$ for targeted Muon. Averaged over matrices, the top arm lifts about 1\% to 28\% of the singular directions, and the bottom arm lifts from nearly all of the directions down to about 30\%. In terms of squared Frobenius norm, the lifted band holds 32--96\% of the momentum in the top arm and at most 13\% in the bottom arm. Note that $\tau$ is absolute while the bulk of the spectrum sits at different levels in different matrices, so the same $\tau$ lifts very different fractions of different matrices. In the top arm this compounds: a matrix whose spectrum falls below $\tau$ only gets its raw momentum and effectively stops training, so at the larger thresholds most of the deeper layers are left untouched. Also note that every momentum singular value is far below one, so the targeted region is always amplified, not suppressed.

\begin{figure*}[t]
\centering
\begin{tabular}{ccc}
    \includegraphics[width=0.3\textwidth]{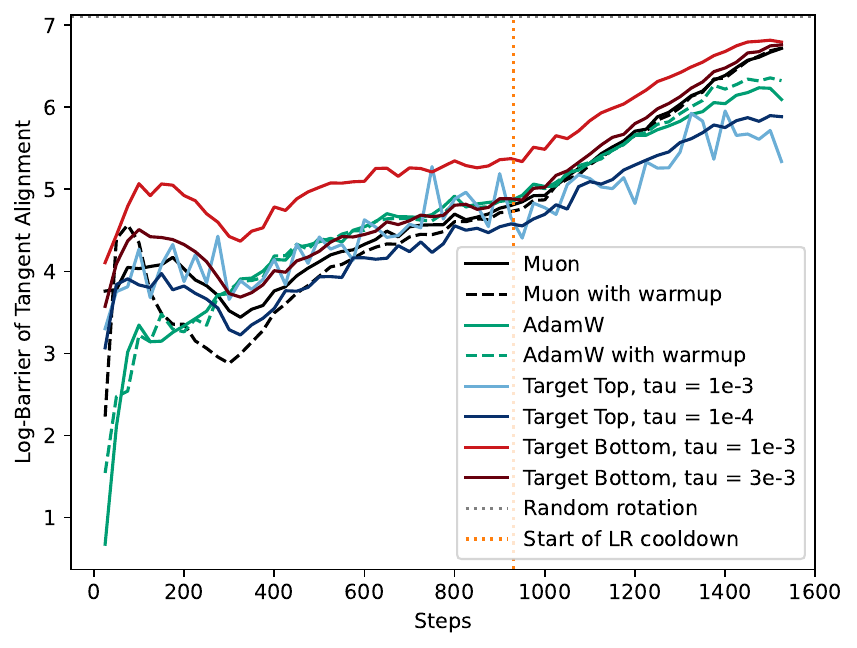} & \includegraphics[width=0.3\textwidth]{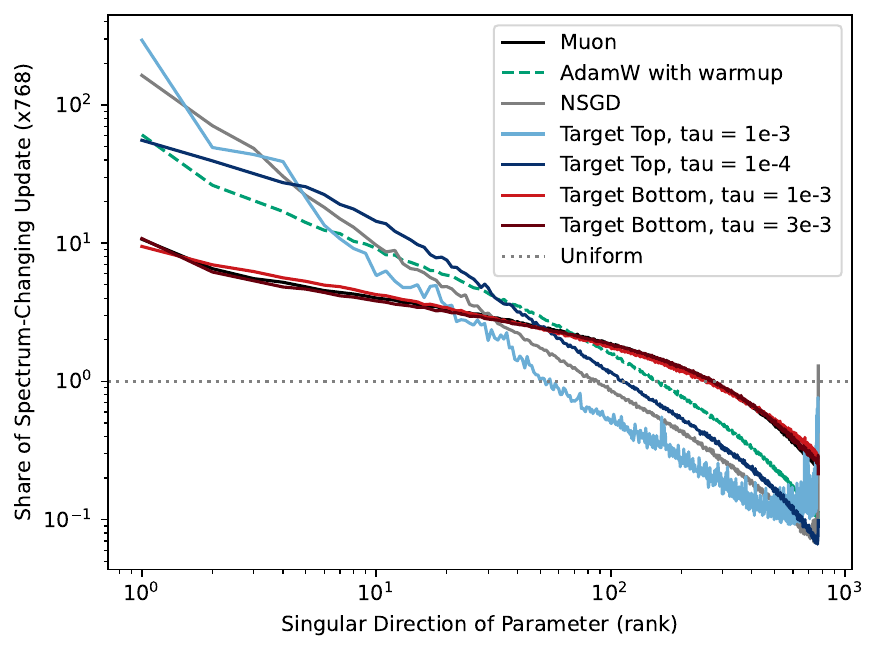} & \includegraphics[width=0.3\textwidth]{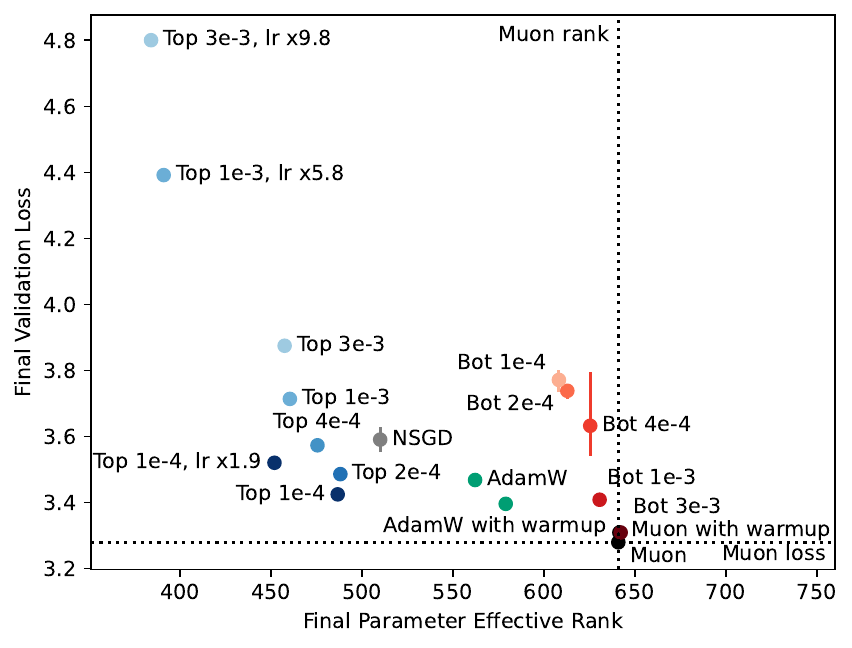} \\
    \parbox[t]{0.3\textwidth}{\centering (a) Tangent alignment} & \parbox[t]{0.3\textwidth}{\centering (b) Spectrum-changing update by singular direction} & \parbox[t]{0.3\textwidth}{\centering (c) Final loss vs.\ parameter rank} \\
\end{tabular}
\vspace{-4mm}\caption{\label{structure}How much of the update changes the spectrum, and where. (a) Log-barrier of the isospectral tangent fraction, averaged over matrices; the grey dotted line is the value for a random rotation, the orange dotted line marks the start of the learning rate cooldown, and dashed lines are the warmup variants. (b) Share of the spectrum-changing part of the update at each singular direction of the parameter, ranked by singular value, averaged over matrices and steps 100--1500; the dotted line is uniform. (c) Final validation loss against final parameter effective rank, including the learning rate controls; the bottom-targeted runs at $\tau\le4\times10^{-4}$ leave their MLP output matrices barely trained. }
\end{figure*}

Validation loss was monitored throughout and remained almost exactly at training loss, so overfitting is not a concern. Every Muon variant drops the loss faster than AdamW over the first hundred steps (Figure~\ref{distloss}(b)), and the lead grows with how much of the spectrum is lifted. Lifting only the top few directions with a high $\tau$ already gets much of Muon's early advantage, but the lead reverses within a few hundred steps, and every top dose ends up behind AdamW with warmup, the smallest ones even behind NSGD (Table~\ref{final_val_loss}). This is not a matter of step size. Raising the learning rate of the top variants only makes them worse and collapses their rank further (Figure~\ref{structure}(c)), while Muon at a tenth of its learning rate loses just 0.05.

Figure~\ref{distloss}(c,d) plots the final loss against $\tau$ and against the fraction of the momentum that is orthogonalized. For both top and bottom targeting, loss falls monotonically with the portion of directions orthogonalized. However, the two arms are not symmetric. Orthogonalizing everything except the few largest singular values of each matrix gets within 0.03 of Muon while orthogonalizing only about 13\% of the momentum matrix's magnitude, while lifting only the top directions remains well short even when they make up 96\% of it, and the smallest top doses do worse than lifting nothing. This answers one part of our original question. Despite being a very small portion of the momentum magnitude, the small singular values are not noise. The model can be usefully trained on them, but only once they are amplified to full scale. 

We use the effective rank of the gradient and update matrix to help explain our optimizer performances. Since the gradient depends only on the surrounding loss landscape and not the optimizer, it tells us which optimizers enter parts of the loss landscape where many directions are available. This happens quickly for Muon and the targeted variants, while AdamW takes a few hundred steps to get there. During learning rate cooldown, this effective rank grows for every optimizer, because the leading singular values of the gradient shrink while its bulk stays put. In the river-valley picture of \citet{wen2024rivervalley}, the component of the gradient along the sharp valley walls dies with the learning rate while the component along the flat floor remains. Gradient rank thus tells us where the iterate sits in the valley, bouncing between the walls or settled on the floor. Meanwhile, the effective rank of the update shows the difference between the different types of targeting, with bottom targeting creating an update shaped like Muon's, while top targeting creates an even lower rank update than AdamW.

Next, we examined the parameter effective ranks across different types of Transformer parameters in Figure~\ref{eranks}(c)--(e). We were able to observe the same results as \citet{wang2025muonoutperformsadamtailend}, who found that AdamW heavily reduces the effective rank of parameters, unlike Muon, giving Muon a richer associative memory and better performance on tail classes. However, top targeted variants fall the most, and their zero-initialized output projections never gain much rank at all. This implies that the rich learning is a direct consequence of participation by the small singular values.

The parameter effective rank depends on the part of the update that changes the spectrum, meaning the part aligned with the parameter's own singular basis, and how concentrated it is in the leading directions of this basis. Figure~\ref{structure}(b) shows this directly: the top targeted and AdamW curves peak at the leading directions, while Muon and the bottom arm stay near uniform. Intuitively, an update that keeps pushing along the leading singular directions of a matrix reinforces what is already there, while a full-spectrum update mostly rotates the basis and leaves the spectrum unchanged. Figure~\ref{structure}(c) plots these against the final loss. Figure~\ref{structure}(a) shows the same thing with the isospectral tangent metric of Section \ref{tangent}. Every optimizer puts more of its update into the parameter's own basis than a random update would, and tangent alignment rises for every optimizer over training, fastest during the cooldown. Among the targeted variants of Muon, the bottom targeted versions show higher alignment than Muon, while the top targeted versions show lower alignment. This is reversed compared to the CIFAR experiments, but it reveals the same underlying phenomenon. In the NanoGPT domain, the targeted region has its magnitude raised, while the remaining regions are relatively suppressed. Therefore, it is simply the top few directions of the gradient which lie within the existing basis of the parameter, and the more you shrink them in the update, the less poorly conditioned the parameter matrices become. Note that structure and performance don't move together: AdamW collapses rank and still performs well, the bottom targeted variants keep Muon's rank without matching its loss, and full-spectrum updates keep the spectrum flat for free. This suggests that keeping the spectrum flat is cheap, but isn't what drives performance.

Our work also suggests an explanation as to why warmup costs Muon but helps AdamW, using the structural plots. AdamW is slow to build structure: its zero-initialized output projections take the whole run to build rank and are still rising at the end (Figure~\ref{eranks}(d)), its early updates lie almost entirely in the parameters' own frame (Figure~\ref{structure}(a)), and its gradients have the lowest effective rank of any optimizer over the first hundred steps, lower than any Muon variant that builds structure at all. An elementwise normalized update builds a matrix one bounded increment at a time, so how well it builds depends on how big the first steps are, and with warmup AdamW ends with higher parameter rank and lower loss. An orthogonalized update writes a full-rank matrix in a single step, so Muon's structure is the same with or without warmup and warmup only throws away full-rate steps.

\clearpage

\section{Conclusion and Future Work}

In this paper, we developed a spectrally targeted variant of the Muon optimizer in order to explore why Muon has so much experimental success. By targeting Muon to act on certain singular values either above or below a certain threshold $\tau,$ we were able to do more specific experiments on Muon's behavior.

From our experiments on CIFAR-10, we concluded that Muon's accuracy improvement requires normalization of all the singular values; by varying $\tau,$ we found that if we normalize singular values, we will measure better evaluation accuracies. However, normalizing only the top singular values yields better tangent alignment and stable weight spectra, hence the primary structural improvements of Muon result from normalizing large singular values, although normalizing smaller ones is important as well. 

From our experiments on NanoGPT, we concluded that the small singular values of the momentum are not noise but directions the model can feasibly train on once they are amplified. Orthogonalizing everything but the few largest singular values nearly matches Muon, while orthogonalizing only the top buys an early lead over AdamW and then falls behind even normalized SGD. Keeping the top from dominating the update is what keeps the parameter spectrum flat; the tangent alignment shows that the top singular directions are the ones that lie outside the tangent space of the isospectral manifold, so they are what changes the spectrum. This is cheap, however, and is not what drives the loss, as AdamW collapses rank and still performs well. AdamW differs from Muon mainly in how slowly it builds structure, because an elementwise normalized update must build a matrix out of many low rank steps, where an orthogonalized update writes a full-rank matrix in a single step. This explains AdamW’s slower start and why warmup helps it but hurts Muon. Finally, the effective rank of the gradient becomes an effective tool to determine position in the loss landscape, and also shows the difference between optimizers in where the early steps take the parameters.

In future work, we could study the effect on larger transformers to see if our observations scale. Also, since matrices spectra sit at different scale,  a threshold relative to the bulk of each spectrum would remove the impact of orthogonalizing a different fraction of directions for each matrix. The radial component of the update also defines an equilibrium weight decay that varies over training and across optimizers, which could be used to create some form of preconditioned weight decay which would prevent change in the spectrum. Given our results, it would also be interesting to consider whether dynamically adapting $\tau$ might improve results, as we could normalize either small or large singular values depending on how far along we are in our training.

\section{\label{appendix}Appendix}
Given $W=U_0\Sigma V_0^\top$ (in unreduced SVD), its isospectral manifold $\mathcal{M}$ is defined as \[\mathcal{M}=\{U\Sigma V^\top\mid U\in O(m),V\in O(n)\}.\] A path $\gamma:(-\varepsilon,\varepsilon)\to\mathcal{M}$ with $\gamma(0)=W$ can be written as $\gamma(t)=U(t)\Sigma V(t)^\top,$ with $U(0)=U_0,$ and $V(0)=V_0.$ We can calculate \begin{align*}\gamma'(0)&=U'(0)\Sigma(V(0))^\top+U(0)\Sigma V'(0)^\top\\&=U'(0)\Sigma V_0^\top+U_0\Sigma V'(0)^\top.\end{align*} A basic result in Lie theory yields that the vector space of possible $U'(0)$ is given by $U(0)A$ for $A\in\mathfrak{so}(m),$ the skew-symmetric $m\times m$ matrices. Similarly, $V'(0)=V(0)B$ for $B\in\mathfrak{so}(n).$ Thus, as $B^\top=-B,$ we see that $\gamma'(0)=U_0(A\Sigma -\Sigma B)V_0^\top,$ so \[T_W\mathcal{M}=\{U_0(A\Sigma -\Sigma B)V_0^\top\mid A\in\mathfrak{so}(m),B\in\mathfrak{so}(n)\}.\]

For simplicity, assume that the diagonal elements of $\Sigma$ are distinct and nonzero, as in the generic case. Given an update matrix $Z\in \mathbb{R}^{m\times n},$ we wish to project $Z$ onto $T_W\mathcal{M}$ with respect to the Frobenius norm, which means that we wish to choose $A\in\mathfrak{so}(m),B\in\mathfrak{so}(n)$ that minimizes \[\Vert Z-U_0(A\Sigma-\Sigma B)V_0^\top\Vert^2_F,\] with the projection given by $U_0(A\Sigma-\Sigma B)V_0^\top.$ By unitary-invariance of the Frobenius norm, this is equivalent to minimizing $\Vert \tilde{Z}-(A\Sigma-\Sigma B)\Vert^2_F,$ where $\tilde{Z}=U_0^\top Z V_0.$ We see that the $ij$th entry of $A\Sigma-\Sigma B$ is $A_{ij}\sigma_j-\sigma_iB_{ij},$ so the diagonal entries must all be zero. However, it can be determined that $A$ and $B$ together have enough degrees of freedom to generate the off-diagonal entries of $\tilde{Z}$, as the singular values are positive and distinct, so we see that the projection of $Z$ onto the tangent space is given by $U_0\tilde{Z}' V_0^\top,$ where the matrix $\tilde{Z}'$ is just $\tilde{Z}$ with the diagonal zeroed out. If some singular values are zero or repeated, the projection computation is more intricate, but this is not relevant for the generic scenario.

Once we calculate the projection, we can use the Frobenius inner product to determine the fraction $\rho(Z;W)$ of the update matrix lies in the tangent space.

\clearpage
\bibliographystyle{icml2025}
\bibliography{references}

@misc{jordan2024muon,
  author       = {Keller Jordan and Yuchen Jin and Vlado Boza and Jiacheng You and
                  Franz Cesista and Laker Newhouse and Jeremy Bernstein},
  title        = {Muon: An optimizer for hidden layers in neural networks},
  year         = {2024},
  url          = {https://kellerjordan.github.io/posts/muon/}
}

@misc{liu2025muonscalablellmtraining,
      title={Muon is Scalable for LLM Training}, 
      author={Jingyuan Liu and Jianlin Su and Xingcheng Yao and Zhejun Jiang and Guokun Lai and Yulun Du and Yidao Qin and Weixin Xu and Enzhe Lu and Junjie Yan and Yanru Chen and Huabin Zheng and Yibo Liu and Shaowei Liu and Bohong Yin and Weiran He and Han Zhu and Yuzhi Wang and Jianzhou Wang and Mengnan Dong and Zheng Zhang and Yongsheng Kang and Hao Zhang and Xinran Xu and Yutao Zhang and Yuxin Wu and Xinyu Zhou and Zhilin Yang},
      year={2025},
      eprint={2502.16982},
      archivePrefix={arXiv},
      primaryClass={cs.LG},
      url={https://arxiv.org/abs/2502.16982}, 
}

@misc{pennington2017resurrectingsigmoiddeeplearning,
      title={Resurrecting the sigmoid in deep learning through dynamical isometry: theory and practice}, 
      author={Jeffrey Pennington and Samuel S. Schoenholz and Surya Ganguli},
      year={2017},
      eprint={1711.04735},
      archivePrefix={arXiv},
      primaryClass={cs.LG},
      url={https://arxiv.org/abs/1711.04735}, 
}

@misc{amsel2026polarexpressoptimalmatrix,
      title={The Polar Express: Optimal Matrix Sign Methods and Their Application to the Muon Algorithm}, 
      author={Noah Amsel and David Persson and Christopher Musco and Robert M. Gower},
      year={2026},
      eprint={2505.16932},
      archivePrefix={arXiv},
      primaryClass={cs.LG},
      url={https://arxiv.org/abs/2505.16932}, 
}

@misc{li2025normuonmakingmuonefficient,
      title={NorMuon: Making Muon more efficient and scalable}, 
      author={Zichong Li and Liming Liu and Chen Liang and Weizhu Chen and Tuo Zhao},
      year={2025},
      eprint={2510.05491},
      archivePrefix={arXiv},
      primaryClass={cs.LG},
      url={https://arxiv.org/abs/2510.05491}, 
}

@article{dewulf2026aurora,
  title   = {Aurora: A Leverage-Aware Optimizer for Rectangular Matrices},
  author  = {Dewulf, Alec and Pai, Dhruv and Yang, Li and Zhang, Ashley
             and Keigwin, Ben},
  year    = {2026},
  url     = {https://tilderesearch.com/blog/aurora}
}

@misc{yukhimchuk2026gradientclippingvectornorms,
      title={Gradient Clipping Beyond Vector Norms: A Spectral Approach for Matrix-Valued Parameters}, 
      author={Alexander Yukhimchuk and Mladen Kolar and Martin Takáč and Sayantan Choudhury},
      year={2026},
      eprint={2605.11838},
      archivePrefix={arXiv},
      primaryClass={cs.LG},
      url={https://arxiv.org/abs/2605.11838}, 
}

@misc{yi2026muconclippedmuonupdates,
      title={MuCon: Clipped Muon Updates for LLM Training}, 
      author={Albert Yi},
      year={2026},
      eprint={2605.26459},
      archivePrefix={arXiv},
      primaryClass={cs.LG},
      url={https://arxiv.org/abs/2605.26459}, 
}

@misc{karpathy2022nanogpt,
  author = {Andrej Karpathy},
  title = {nanoGPT},
  year = {2022},
  publisher = {GitHub},
  journal = {GitHub repository},
  howpublished = {\url{https://github.com/karpathy/nanoGPT}}
}

@techreport{krizhevsky2009learning,
  title={Learning multiple layers of features from tiny images},
  author={Krizhevsky, Alex},
  institution={University of Toronto},
  year={2009}
}

@article{jordan202494,
  title={94\% on CIFAR-10 in 3.29 Seconds on a Single GPU},
  author={Jordan, Keller},
  journal={arXiv preprint arXiv:2404.00498},
  year={2024}
}

@INPROCEEDINGS{7098875,
  author={Roy, Olivier and Vetterli, Martin},
  booktitle={2007 15th European Signal Processing Conference}, 
  title={The effective rank: A measure of effective dimensionality}, 
  year={2007},
  volume={},
  number={},
  pages={606-610},
  doi={}}

@misc{wang2025muonoutperformsadamtailend,
      title={Muon Outperforms Adam in Tail-End Associative Memory Learning}, 
      author={Shuche Wang and Fengzhuo Zhang and Jiaxiang Li and Cunxiao Du and Chao Du and Tianyu Pang and Zhuoran Yang and Mingyi Hong and Vincent Y. F. Tan},
      year={2025},
      eprint={2509.26030},
      archivePrefix={arXiv},
      primaryClass={cs.LG},
      url={https://arxiv.org/abs/2509.26030}, 
}

@misc{ai2025practicalefficiencymuonpretraining,
      title={Practical Efficiency of Muon for Pretraining}, 
      author={Ishaan Shah and Anthony M. Polloreno and Karl Stratos and Philip Monk and Adarsh Chaluvaraju and Andrew Hojel and Andrew Ma and Anil Thomas and Ashish Tanwer and Darsh J Shah and Khoi Nguyen and Kurt Smith and Michael Callahan and Michael Pust and Mohit Parmar and Peter Rushton and Platon Mazarakis and Ritvik Kapila and Saurabh Srivastava and Somanshu Singla and Tim Romanski and Yash Vanjani and Ashish Vaswani},
      year={2025},
      eprint={2505.02222},
      archivePrefix={arXiv},
      primaryClass={cs.LG},
      url={https://arxiv.org/abs/2505.02222}, 
}

@misc{penedo2024finewebdatasetsdecantingweb,
      title={The FineWeb Datasets: Decanting the Web for the Finest Text Data at Scale}, 
      author={Guilherme Penedo and Hynek Kydlíček and Loubna Ben allal and Anton Lozhkov and Margaret Mitchell and Colin Raffel and Leandro Von Werra and Thomas Wolf},
      year={2024},
      eprint={2406.17557},
      archivePrefix={arXiv},
      primaryClass={cs.CL},
      url={https://arxiv.org/abs/2406.17557}, 
}

@article{DBLP:journals/corr/RonnebergerFB15,
  author       = {Olaf Ronneberger and
                  Philipp Fischer and
                  Thomas Brox},
  title        = {U-Net: Convolutional Networks for Biomedical Image Segmentation},
  journal      = {CoRR},
  volume       = {abs/1505.04597},
  year         = {2015},
  url          = {http://arxiv.org/abs/1505.04597},
  eprinttype   = {arXiv},
  eprint       = {1505.04597},
  bibsource    = {dblp computer science bibliography, https://dblp.org}
}

@misc{jordan2024moddednanogpt,
  author = {Keller Jordan and Jeremy Bernstein and Brendan Rappazzo and others},
  title = {modded-nanogpt: Speedrunning the NanoGPT baseline},
  year = {2024},
  howpublished = {\url{https://github.com/KellerJordan/modded-nanogpt}}
}

@misc{freon2026,
  title={Muon is Not That Special: Random or Inverted Spectra Work Just as Well},
  author={Zakhar Shumaylov and Nathaël Da Costa and Peter Zaika and Bálint Mucsányi and Alex Massucco and Yoav Gelberg and Carola-Bibiane Schönlieb and Yarin Gal and Philipp Hennig},
  year={2026}, eprint={2605.11181}, archivePrefix={arXiv}, primaryClass={cs.LG}
}

@misc{schattenp2026,
  title={From SGD to Muon: Adaptive Optimization via Schatten-p Norms},
  author={Thomas Massena and Corentin Friedrich and Mathieu Serrurier},
  year={2026}, eprint={2605.19781}, archivePrefix={arXiv}, primaryClass={cs.LG}
}

@misc{muonp2026,
  title={Muon$^p$: Muon with Fractional Spectral Powers},
  author={Yihe Dong and Will Sawin},
  year={2026}, eprint={2606.13867}, archivePrefix={arXiv}, primaryClass={cs.LG}
}

@misc{dynmuon2026,
  title={DynMuon: A Dynamic Spectral Shaping View of Muon},
  author={Fangzhou Wu and Rikhav Shah and Sandeep Silwal and Qiuyi Zhang},
  year={2026}, eprint={2605.17109}, archivePrefix={arXiv}, primaryClass={cs.LG}
}

@misc{pion_highpass2026,
  title={Rethinking Muon Beyond Pretraining: Spectral Failures and High-Pass Remedies for VLA and RLVR},
  author={Chongyu Fan and Gaowen Liu and Mingyi Hong and Ramana Rao Kompella and Sijia Liu},
  year={2026}, eprint={2605.19282}, archivePrefix={arXiv}, primaryClass={cs.LG}
}

@misc{pion_preserving2026,
  title={Pion: A Spectrum-Preserving Optimizer via Orthogonal Equivalence Transformation},
  author={Kexuan Shi and Hanxuan Li and Zeju Qiu and Yandong Wen and Simon Buchholz and Weiyang Liu},
  year={2026}, eprint={2605.12492}, archivePrefix={arXiv}, primaryClass={cs.LG}
}

@misc{wen2024rivervalley,
  title={Understanding Warmup-Stable-Decay Learning Rates: A River Valley Loss Landscape Perspective},
  author={Kaiyue Wen and Zhiyuan Li and Jason Wang and David Hall and Percy Liang and Tengyu Ma},
  year={2024}, eprint={2410.05192}, archivePrefix={arXiv}, primaryClass={cs.LG}
}
\end{document}